\documentclass[11pt]{article}
\usepackage[final]{acl}
\usepackage{times}
\usepackage{latexsym}
\usepackage[T1]{fontenc}
\usepackage[utf8]{inputenc}
\usepackage{url}
\usepackage{booktabs}
\usepackage{amsfonts}
\usepackage{amsmath}
\usepackage{amssymb}
\usepackage{amsthm}
\usepackage{nicefrac}
\usepackage{microtype}
\usepackage{inconsolata}
\usepackage{xcolor}
\usepackage{graphicx}
\graphicspath{{figures/}{block-granularity/figures/}}
\usepackage{subcaption}
\usepackage{algorithm}
\usepackage{algorithmic}
\usepackage{mathtools}
\usepackage{multirow}
\usepackage{enumitem}
\usepackage{placeins}
\usepackage{float}
\usepackage{tikz}
\usetikzlibrary{arrows.meta,positioning,fit,backgrounds,calc,decorations.pathreplacing}

\title{Per-Matrix Optimality Is Not Enough:\\Three-Level Optimization for Low-Rank LLM Compression}

\renewcommand{\thefootnote}{\fnsymbol{footnote}}
\newcommand{\aff}[1]{\textsuperscript{\normalfont #1}}
\author{
  Huicheng Zhang\aff{1}\footnotemark[1] \quad
  Xiyao Feng\aff{1}\footnotemark[1] \quad
  Ze-Tong Li\aff{1} \quad
  Chengkai Zhu\aff{1,2} \\
  \bfseries Xiao Shi\aff{1}\footnotemark[2] \quad
  Xiwei Pan\aff{1} \quad
  Jinguo Liu\aff{1}\footnotemark[2] \quad
  Ge Bai\aff{1} \quad
  Xin Wang\aff{1}\footnotemark[2] \\[2pt]
  \texttt{\{xiaoshi,jinguoliu,felixxinwang\}@hkust-gz.edu.cn} \\[2pt]
  \aff{1}Hong Kong University of Science and Technology (Guangzhou), Guangdong 511453, China \\
  \aff{2}QudeLeap Research, Shanghai 200030, China
}
\begin{document}

\maketitle
\footnotetext[1]{\, Equal contribution.}
\footnotetext[2]{\, Corresponding author.}
\renewcommand{\thefootnote}{\arabic{footnote}}

\begin{abstract}
Per-matrix singular value decomposition (SVD) truncation is
Eckart--Young optimal in the whitened Frobenius norm, but errors
from independently compressed matrices compound through the
block's nonlinear forward pass.
Inspired in part by hierarchical variational optimization in quantum
many-body methods, we introduce a three-level chain that widens optimization scope from
individual matrices to Transformer blocks to the full model:
whitened SVD~(L1), block-level joint optimization~(L2), and
end-to-end language-modeling loss refinement~(L3), all from
256 calibration sequences, with no instruction or recovery data. 
On LLaMA-7B at 60\% compression, the chain reduces WikiText-2 perplexity from 42.1 to 19.1 to 11.4.
The block-level stage acts as a regularizer: skipping it worsens Penn Treebank (PTB) perplexity by 24 points, a gap that additional end-to-end training did not close in our experiments.
Perplexity gains hold across 20--80\% compression, five
architectures up to 13B parameters, and both in-distribution and out-of-distribution benchmarks, though the cross-architecture rows use architecture-specific configurations and the ratio sweep was not run under one common protocol.
With more calibration data, skipping the block-level stage becomes competitive, revealing an offline compute--data trade-off. We therefore claim improvements only in perplexity and compression fidelity; downstream accuracy remains well below the dense model.
\end{abstract}

\section{Introduction}
\label{sec:intro}

\begin{figure*}[t]
\centering
\includegraphics[width=\textwidth]{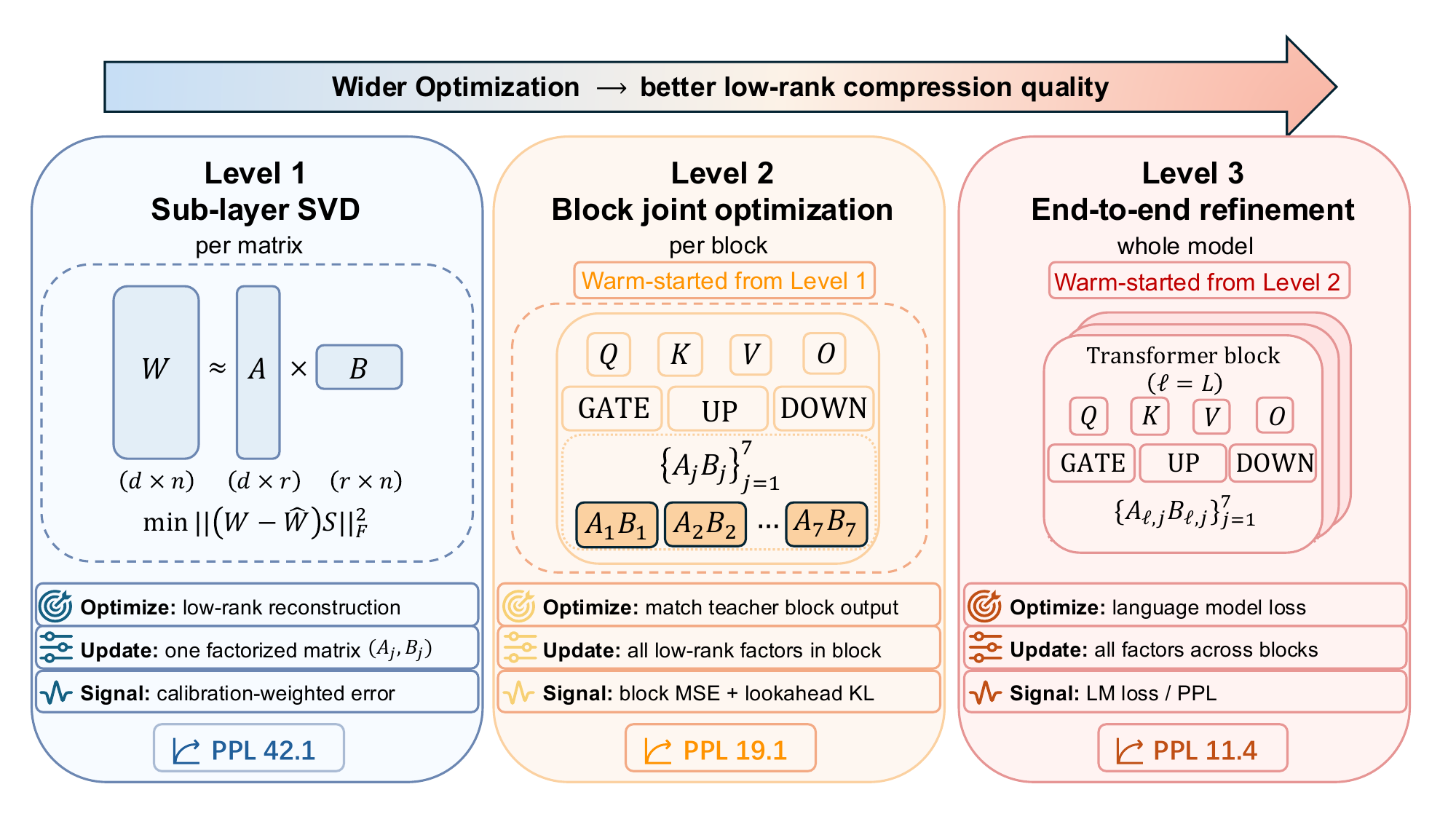}
\caption{Three-level compression chain.
Each level widens the optimization scope and warm-starts from the previous solution using the same calibration budget. 
Numbers: LLaMA-7B WikiText-2 PPL at $\rho{=}0.6$. Each level reduces perplexity by roughly $2\times$.}
\label{fig:hierarchy}
\end{figure*}


Large language models (LLMs) require significant memory, limiting deployment on consumer-grade hardware. Post-training compression reduces this cost through weight quantization~\cite{frantar2022gptq,lin2024awq}, pruning~\cite{frantar2023sparsegpt,wei2024structured}, or low-rank decomposition. Singular value decomposition (SVD)-based low-rank compression preserves the original Transformer structure and uses only standard dense matrix multiplications with no specialized kernels.
Among the earliest applications to neural network compression, \citet{sainath2013low} factorized the final weight layer of deep neural network (DNN) acoustic models; subsequent work extended low-rank approximations to convolutional neural network (CNN) filters~\cite{denton2014exploiting,jaderberg2014speeding}. Truncating the SVD of each LLM weight matrix directly is a natural baseline, but this data-agnostic objective ignores which input directions dominate the projection's output error. Activation-aware methods address this mismatch: ASVD~\cite{yuan2023asvd} rescales weight columns by activation-channel magnitude before SVD, while SVD-LLM~\cite{wang2025svd} whitens calibration activations so that singular-value truncation directly corresponds to reconstruction loss. Follow-up work refines per-matrix objectives~\cite{wang2025dobi}, allocates rank budgets across weight matrices~\cite{xv2025ara,qi2026swift,ding2025dipsvd,abbasi2026zero}, and compensates accumulated error~\cite{hu2026saes}. To widen the scope beyond individual matrices, LiLlama~\cite{sy2024lillama} performs SVD-initialized per-block distillation with 13\,M calibration tokens, but remains local in scope.

All these methods optimize a local reconstruction surrogate for each weight matrix independently.
Activation-whitened SVD already achieves the Eckart--Young optimum~\cite{eckart1936approximation} for this per-matrix objective, yet on LLaMA-7B at per-matrix removal ratio $\rho{=}0.6$ it yields a WikiText-2 perplexity of 42.1, far from the dense model's 5.68.
The missing error enters at two scales: within each Transformer block, truncation errors from the seven attention and multilayer perceptron (MLP) projections interact through residual connections and nonlinearities; across blocks, these perturbations accumulate along the residual stream before reaching the language-modeling loss.
Neither scale is visible to a per-matrix or per-block objective.
Low-Rank Adaptation (LoRA) fine-tuning~\cite{hu2021lora} on external data such as 50K Alpaca samples can recover perplexity, but it changes the setting from self-contained post-training compression to data-dependent recovery.
Post-training low-rank compression, therefore, needs a broader optimization scope, one that captures interactions within and across Transformer blocks using only the calibration data already required for compression.

In this paper, we argue that the unit of post-training low-rank compression should not be the weight matrix but the computation
graph over which truncation errors interact.
Starting from whitened per-matrix SVD, we optimize the same low-rank factors at progressively larger scopes, first through each Transformer block's nonlinear forward pass, then through the full model's language-modeling (LM) loss, using only the original 256 calibration sequences and adding no inference-time parameters. WikiText-2 validation is used only for L3 early stopping (Figure~\ref{fig:hierarchy}).
On LLaMA-7B at 60\% compression, WikiText-2 perplexity drops from
42.1 after per-matrix SVD to 19.1 after block-level optimization
and 11.4 after full-model optimization, lower than SVD-LLM +
Sequential LoRA~(15.0) despite using 256 samples rather than
50K~\cite{wang2025svd}.
The same pattern holds across 20--80\% compression
(Tables~\ref{tab:svd-comparison} and~\ref{tab:ratios}), on held-out
Penn Treebank~\cite{marcus-etal-1993-building} (PTB) and C4~\cite{JMLR:v21:20-074}, and on five architectures 
(Table~\ref{tab:cross-arch}).
The block-level stage also acts as a regularizer: skipping it worsens out-of-distribution (OOD) perplexity by 24 points, a gap that additional end-to-end training did not close in our experiments
(\S\ref{sec:ablations}).
Our specific methodological contribution is the block objective of
\S\ref{sec:method}---cross-frontier reconstruction with a next-block
vocabulary lookahead---together with the finding that widening
optimization scope, rather than any single decomposition, closes most
of the gap; we claim no novelty for whitened SVD, generic block-level
recovery, or the LM loss on their own.

\section{Background and Related Work}
\label{sec:related}

\paragraph{Per-Matrix SVD Compression.}
Post-training LLM compression spans
quantization~\cite{frantar2022gptq,lin2024awq},
structured pruning~\cite{frantar2023sparsegpt,wei2024structured},
and low-rank factorization. We focus on the last.
Per-matrix SVD methods vary along three axes---\emph{objective
design}, \emph{error compensation}, and \emph{rank
allocation}---but all factorize each weight matrix independently.
On the objective side,
SVD-LLM~\cite{wang2025svd} whitens activations so that the
Frobenius surrogate reflects input statistics.
ASVD~\cite{yuan2023asvd} scales weight columns by activation
magnitudes before decomposition, and Dobi-SVD~\cite{wang2025dobi} makes truncation
differentiable for gradient-based tuning.
For error compensation, SAES-SVD~\cite{hu2026saes} adaptively
suppresses the error that accumulates when sub-layers are compressed
sequentially.
Zero-Sum SVD~\cite{abbasi2026zero} selects singular components
globally via a zero-sum pruning rule, coupling rank decisions
across sub-layers but leaving each matrix's factorization fixed.
For rank allocation, ARA~\cite{xv2025ara} learns per-module rank budgets,
Swift-SVD~\cite{qi2026swift} obtains per-matrix solutions in closed
form and selects allocations via grid search, and
DipSVD~\cite{ding2025dipsvd} uses channel-weighted whitening and
allocates ranks via Fisher sensitivity and effective rank.
All these methods improve the per-matrix factorization or its rank
budget, but do not jointly optimize across sub-layers within a block.

\paragraph{Beyond Per-Matrix Scope.}
Two lines of work widen the optimization scope beyond the individual matrices, but both remain limited.
On the compression side,
LiLlama~\cite{sy2024lillama} performs SVD-initialized per-block
distillation with a hybrid $\ell_1${+}cosine loss and 13\,M
calibration tokens, and
ERC-SVD~\cite{bai2025ressvd} mitigates cross-layer error
accumulation by leaving early layers uncompressed and applying
residual SVD compensation to the rest. 
AA-SVD~\cite{sinha2026aa} anchors each compressed layer's output to the corresponding original output while adapting to input distribution shifts, then refines each Transformer block end-to-end to minimize block-level output distortion. 
Both still minimize a local reconstruction loss (per-block or
per-matrix) rather than the end-to-end LM loss.
On the recovery side, LoRA
fine-tuning~\cite{hu2021lora} on external data such as
Alpaca can restore perplexity, but adds
a data dependency absent from the compression step.
Our block-level objective is related to knowledge
distillation~\cite{hinton2015distilling}, but operates on
compressed computation graphs under a calibration-only budget.
Few works systematically compare per-matrix, block, and full-model optimization scopes under a fixed calibration budget, and none study the block stage as a regularizer for end-to-end refinement.

Table~\ref{tab:relwork} locates our block stage among these methods.
AA-SVD overlaps with our cross-frontier reconstruction in anchoring
original outputs under shifted inputs; what distinguishes our stage
is the next-block vocabulary lookahead, followed by
factor-constrained LM-loss refinement under the same small
calibration contract.
We therefore present the per-matrix/block/full-model scope
comparison as the broader empirical finding and cross-frontier plus
lookahead (CF+LA) as the specific method contribution, claiming no
novelty for whitened SVD, for generic block-level recovery, for the
LM loss, or for cross-frontier anchoring alone.
Our matched full-versus-skip controls support the complete block
stage as a bridge to end-to-end refinement; they do not isolate the
individual contributions of cross-frontier reconstruction and
lookahead, nor do they establish cross-architecture superiority over
these methods.

\begin{table}[t]
\centering
\footnotesize
\setlength{\tabcolsep}{2pt}
\begin{tabular}{@{}lp{3.0cm}cc@{}}
\toprule
Method & Block/local recovery signal & LA & LM \\
\midrule
LiLlama        & local feature distillation & no & no$^\ast$ \\
AA-SVD         & original-output anchoring under shifted inputs & no & no \\
ERC-SVD        & residual compensation, selective compression & no & no \\
\textbf{Ours}  & cross-frontier block reconstruction & \textbf{yes} & \textbf{yes} \\
\bottomrule
\multicolumn{4}{@{}l}{\scriptsize $^\ast$Not part of its core recovery procedure.}
\end{tabular}
\caption{Where our block-level stage sits among prior block and
per-layer recovery methods. ``LA'' is whether the block objective penalizes a vocabulary-space
shift measured through the following block (next-block lookahead);
``LM'' is whether the low-rank factors are subsequently refined
against the end-to-end language-modeling loss.}
\label{tab:relwork}
\end{table}

\section{Three-Level Optimization}
\label{sec:theory}

When each weight matrix is compressed independently, the resulting
errors interact through the Transformer block's nonlinear forward
pass in ways that per-matrix objectives cannot anticipate.
Across blocks, these errors compound further, progressively
degrading end-to-end performance.
Our three-level chain counters this by widening the optimization
scope at each stage.
L1 performs whitened SVD on individual weights, serving as a strong
per-matrix baseline (\S\ref{sec:l1}).
L2 jointly optimizes all factors within each block to capture
cross-sub-layer error interactions (\S\ref{sec:method}).
L3 extends optimization to the full model via end-to-end LM-loss
refinement, correcting cross-block error accumulation
(\S\ref{sec:level3}).
Each level warm-starts from the previous solution, and all three use only the calibration set without external data.

In the following text, let $\mathcal{D}_{\mathrm{cal}}=\{x_i\}_{i=1}^{N}$ be a calibration set of $N$ token sequences shared by all three levels, where each
$x_i=(x_i^{(1)},\ldots,x_i^{(T_i)})$ is a token sequence of length $T_i$. Consider a model with $L$ Transformer blocks. Let $f_\ell$, $\ell\in\{0,\dots,L-1\}$, denote the $\ell$-th Transformer block with weight matrices $W_{\ell,j} \in \mathbb{R}^{d_j \times n_j}$, $j\in \mathcal{S}$. Here, without loss of generality, let the index set $\mathcal{S}:=\{q,k,v,o,\mathrm{gate},\mathrm{up},\mathrm{down}\}$ such that $W_{\ell,j}$ represent the weight matrices of $Q$-, $K$-, $V$-, and $O$-projections of the
attention mechanism, and the gate, up, and down projections of the
MLP, respectively. 

Running the dense model on $\mathcal{D}_{\mathrm{cal}}$ produces
hidden-state matrices $h_\ell \in \mathbb{R}^{d \times T}$ with forward rule $h_{\ell+1} = f_\ell(h_\ell; \{W_{\ell,j}\})$, where $T = \sum_{i=1}^{N} T_i$ is the total number of tokens, and $h_0$ is the token embeddings of the calibration sequences.

\subsection{Per-Matrix Whitened SVD (L1)}
\label{sec:l1}

L1 follows the whitened SVD procedure of
SVD-LLM~\cite{wang2025svd}. Given the removal ratio
  $\rho \in (0,1)$ indicating the fraction of parameters removed per weight matrix, the L1 finds the low-rank approximation of each weight matrix as $W_{\ell,j} \approx_{\mathcal{D}_{\mathrm{cal}}} A_{\ell,j} B_{\ell,j}$ based on the calibration data $\mathcal{D}_{\mathrm{cal}}$, where $A_{\ell,j} \in \mathbb{R}^{d_j \times r_j}$,
$B_{\ell,j} \in \mathbb{R}^{r_j \times n_j}$, and the rank
$r_j = \lfloor (1{-}\rho)\,d_j n_j / (d_j{+}n_j) \rfloor$.

Specifically, for the weight matrix $W_{\ell,j}$ with activation $X_{\ell,j}$, L1 solves
\begin{equation}
    \min_{A_{\ell,j}, B_{\ell,j}}\|(W_{\ell,j}{-}A_{\ell,j}B_{\ell,j})X_{\ell,j}\|_F^2.
\end{equation}
From \cite{wang2025svd,eckart1936approximation}, the analytical solution is given by
\begin{equation}
    A_{\ell,j} = U_{r_{j}} \sqrt{\Sigma_{r_j}},~ B_{\ell,j} = \sqrt{\Sigma_{r_j}} V_{r_{j}} S^{-1},
\end{equation}
where $U_{r_{j}} \Sigma_{r_{j}} V_{r_{j}}^\top$ is the rank-$r_{j}$ truncated SVD of $W_{\ell,j}S_{\ell,j}$ and $S_{\ell,j}$ is the lower Cholesky factor of the activation covariance $XX^\top$. 


Although L1 attains the theoretical optimum in terms of the activation‑aware Frobenius distance for each weight matrix independently, it neglects the cross‑interactions among the attention projections and the MLP projections.
As these per‑matrix errors propagate through the block’s nonlinear forward pass, they accumulate in ways a per‑matrix objective cannot capture. Hence, even a globally optimal per‑matrix approximation still underperforms the dense model, motivating our block‑level joint optimization in L2.

\subsection{Block-Level Joint Optimization (L2)}
\label{sec:method}

L2 jointly optimizes all low‑rank factors within a block (Algorithm~\ref{alg:block}), allowing approximation errors to be redistributed across attention and MLP components -- a form of error compensation that is infeasible under per‑matrix optimization.

The block loss is defined by
\begin{align}
\label{eq:block-loss-overview}
\mathcal{L}_\ell = \mathcal{L}_{\mathrm{rec}} + \lambda_{\mathrm{la}}\,\mathcal{L}_{\mathrm{la}},
\end{align}
where $\mathcal{L}_{\mathrm{rec}}$ measures the accumulated reconstruction error
between the compressed and original blocks, and
$\mathcal{L}_{\mathrm{la}}$ is a look-ahead term that
penalizes errors amplified by the next block, with
coefficient $\lambda_{\mathrm{la}} > 0$.

Let $\hat{h}_{\ell+1} = f_\ell(\hat{h}_{\ell}; \{A_{\ell,j} B_{\ell,j}\})$ denote the compressed hidden-state matrices with initialization $\hat{h}_0 = h_0$. Then, the reconstruction error is defined as the Euclidean distance
\begin{align}
\label{eq:block-obj}
\mathcal{L}_{\mathrm{rec}} =
\bigl\| h_{\ell+1} -
\hat{h}_{\ell+1} \bigr\|_F^2.
\end{align}
Minimizing $\mathcal{L}_{\mathrm{rec}}$ requires the compressed block
to reproduce the original output $h_{\ell+1}$ from a degraded input
$\hat{h}_\ell$, compensating for drift accumulated across preceding
blocks. However, this objective cannot capture the amplification of these errors by downstream blocks.

To capture the downstream impact of compressing block $\ell$, we
propagate the hidden states through the frozen next block
$f_{\ell+1}$ and introduce a per-block probe matrix
$P_{\ell+1} \in \mathbb{R}^{|V| \times d}$
that projects the intermediate output of the $\ell+1$-th block to the vocabulary space,
acting as a temporary training-only surrogate for the remaining layers and the final vocabulary projection $W_{\text{head}}$. Specifically, $P_\ell$ is trained on the calibration set so that $\mathrm{softmax}(P_\ell \bar{h}_{\ell+1}/ \tau )$ approximates the token
distribution $\mathrm{softmax}(W_{\text{head}} \bar{h}_{L} / \tau)$
produced by the full model, where $\bar{\cdot} := \mathrm{RMSNorm(\cdot)}$ and $\tau > 0$ is a temperature hyperparameter. Since columns of both quantities lie on the probability simplex after softmax normalization, their discrepancy can be measured by the Kullback--Leibler (KL) divergence. We train each probe by minimizing 
\begin{equation}\label{eq:training_P}
    \begin{aligned}
        \mathcal{L}_{P_{\ell}} = \sum_{i=1}^{T} D_{\mathrm{KL}}\!(&
\mathrm{softmax}(W_{\text{head}} \bar{h}_{L}^{(i)} / \tau) \,\|\,\\
&\mathrm{softmax}({P_\ell \bar{h}_{\ell+1}^{(i)}}/{\tau})),
    \end{aligned}
\end{equation}
where ${h}^{(i)}$ represents the $i$-th column of $h$.
With $P_\ell$ fixed, we define the vocabulary distributions
\begin{align}
p_{\ell+1} = \mathrm{softmax}\!\bigl(P_{\ell+1}\,
\bar{f}_{\ell+1}(h_{\ell+1}) \,/\, \tau\bigr),\\
\hat{p}_{\ell+1} = \mathrm{softmax}\!\bigl(P_{\ell+1}\,
\bar{f}_{\ell+1}(\hat{h}_{\ell+1}) \,/\, \tau\bigr),
\end{align}
where $\bar{f}_{\ell+1}:=\mathrm{RMSNorm}\cdot{f}_{\ell+1}$, and weights are omitted.
Then, the look-ahead loss is given by
\begin{align}
\label{eq:lookahead}
\mathcal{L}_{\mathrm{la}} = \sum_{i=1}^{T}
D_{\mathrm{KL}}\!\left( p_{\ell+1}^{(i)} \,\|\, \hat{p}_{\ell+1}^{(i)} \right)
\end{align}
which measures the next-block distributional shift of token prediction induced by the compression error.
For the final block ($\ell=L{-}1$), no subsequent block is available for look-ahead. We set $\lambda_{\mathrm{la}}=0$ and optimize $\mathcal{L}_{\mathrm{rec}}$ only.
The look-ahead captures the sensitivity of the immediate next block, but multi-block error cascades remain invisible to L2 and motivate the end-to-end L3 stage.

\begin{algorithm}[t]
\caption{Block-Level Joint Optimization (L2).}
\label{alg:block}
\begin{algorithmic}[1]
\REQUIRE Original weights $\{W_{\ell,j}\}_{\ell=0,j=1}^{L-1,\,7}$, L1 results $\{\{A_{\ell,j}, B_{\ell,j}\}_{j\in\mathcal{S}}\}_{\ell=0}^{L-1}$,
calibration dataset $\mathcal{D}_{\mathrm{cal}}$, look-ahead coefficient $\lambda_{\mathrm{la}}$.
\STATE $\{h_0,\dots,h_L\} \leftarrow $ collect hidden states on $\mathcal{D}_{\mathrm{cal}}$
\STATE $\hat{h}_0 \leftarrow h_0$
\FOR{$\ell = 0$ to $L-1$}
    \STATE $P_{\ell+1} \leftarrow \arg\min \mathcal{L}_{P_{\ell+1}}$
    \STATE $\{A_{\ell,j},B_{\ell,j}\}_{j\in\mathcal{S}} \leftarrow \arg\min \mathcal{L}_{\ell}$
    \STATE $\hat{h}_{\ell+1} \leftarrow f_\ell(\hat{h}_{\ell}; \{A_{\ell,j} B_{\ell,j}\})$
\ENDFOR
\RETURN optimized $\{\{A_{\ell,j}, B_{\ell,j}\}_{j\in\mathcal{S}}\}_{\ell=0}^{L-1}$.
\end{algorithmic}
\end{algorithm}


\subsection{End-to-End LM-Loss Refinement (L3)}
\label{sec:level3}

L2 widens optimization scope from individual matrices to blocks, but processes blocks sequentially with detached gradients, so cross-block error accumulation is not captured.
L3 completes the progression to the full model by minimizing the
standard autoregressive next-token prediction loss over
$\mathcal{D}_{\mathrm{cal}}$.
The low-rank factors $\{A_{\ell,j},B_{\ell,j}\}$ are warm-started from L2 and
updated end-to-end. The objective is the standard next-token cross-entropy
\begin{equation}
\label{eq:l3-loss}
\mathcal{L}_{\mathrm{L3}} =
\!-\!{\sum_{i=1}^{N}\sum_{t=1}^{T_i}
\log p\bigl(x_i^{(t+1)} \mid x_i^{(1)},...,x_i^{(t)}\bigr)}
,
\end{equation}
where $p\bigl(x_i^{(t+1)} \mid x_i^{(1)},...,x_i^{(t)}\bigr)$ is the compressed model's
next-token probability computed with all weights $W_{\ell,j}$ replaced by
their low-rank factors $A_{\ell,j} B_{\ell,j}$ as the trainable variables.
Unlike SVD-LLM's Sequential LoRA, which freezes the SVD factors
and trains additional weight matrices on external samples, L3 optimizes
the existing factors without additional parameters or data.

\begin{table*}[t]
\centering
\footnotesize
\setlength{\tabcolsep}{3pt}
\begin{tabular}{@{} cl rrr ccccccc c @{}}
\toprule
  & & \multicolumn{3}{c}{Perplexity $\downarrow$} & \multicolumn{7}{c}{Zero-Shot Accuracy (\%) $\uparrow$} & \\
  \cmidrule(lr){3-5} \cmidrule(lr){6-12}
  $\rho$ & Method & Wiki & PTB & C4 & Openb. & ARC\textsubscript{e} & ARC\textsubscript{c} & WinoG. & HellaS. & PIQA & MathQA & Avg.$\uparrow$ \\
\midrule
--- & \emph{Uncompressed} & \emph{5.68} & \emph{8.35} & \emph{7.34} & \emph{28.0} & \emph{67.0} & \emph{38.0} & \emph{67.0} & \emph{56.0} & \emph{78.0} & \emph{27.0} & \emph{51.6} \\
\midrule
\multirow{6}{*}{0.2}
& ASVD$^\dagger$         & 11.1 & 16.6 & 15.9 & 25.0 & 53.0 & 27.0 & 64.0 & 41.0 & 68.0 & 24.0 & 43.1   \\
& SVD-LLM$^\dagger$      & 7.94 & 16.2 & 15.8 & 22.0 & 58.0 & 29.0 & 63.0 & 43.0 & 69.0 & 24.0 & 44.0   \\
& Dobi-SVD$^{*\dagger}$  & 8.54 & 14.8 & \textbf{10.0} & 26.0 & 59.0 & 31.0 & 66.0 & 44.0 & 70.0 & 23.0 & 45.6 \\
& DipSVD$^*$            & 7.95 & 15.6 & 14.1 & 27.0 & 63.0 & 33.0 & 64.0 & 45.0 & 71.0 & 24.0 & 46.7  \\
& SAES-SVD               & 7.17 & 15.2 & 13.8 & 29.0 & 68.0 & \textbf{36.0} & 65.0 & 45.0 & \textbf{75.0} & 25.0 & 49.0 \\
& \textbf{Ours (L3)}     & \textbf{6.47} & \textbf{14.1} & 13.5 & \textbf{29.8} & \textbf{70.0} & 35.0 & \textbf{66.0} & \textbf{47.6} & 71.7 & \textbf{26.4} & \textbf{49.5}  \\
\midrule
\multirow{6}{*}{0.4}
& ASVD$^\dagger$         & $>$1k & $>$3k & $>$1k & 13.0 & 28.0 & 22.0 & 48.0 & 26.0 & 55.0 & 19.0 & 30.1  \\
& SVD-LLM$^\dagger$      & 13.1 & 63.8 & 49.8 & 19.0 & 42.0 & 25.0 & 58.0 & 33.0 & 60.0 & 21.0 & 36.9  \\
& Dobi-SVD$^{*\dagger}$  & 13.5 & 46.4 & \textbf{23.5} & 22.0 & 41.0 & 27.0 & 58.0 & 34.0 & 61.0 & 23.0 & 38.0  \\
& DipSVD$^*$            & 12.8 & 47.0 & 34.4 & 22.0 & 50.0 & \textbf{30.0} & 61.0 & 36.0 & 64.0 & 22.0 & 40.7 \\
& SAES-SVD               & 10.4 & 45.1 & 32.8 & 23.0 & 50.0 & 29.0 & \textbf{62.0} & 36.0 & 65.0 & 23.0 & 41.1  \\
& \textbf{Ours (L3)}     & \textbf{7.72} & \textbf{31.3} & 25.7 & \textbf{24.2} & \textbf{59.8} & 29.9 & \textbf{62.0} & \textbf{40.4} & \textbf{65.6} & \textbf{24.4} & \textbf{43.8}   \\
\midrule
\multirow{5}{*}{0.6}
& ASVD$^\dagger$         & $>$60k & $>$40k & $>$400k & 12.0 & 26.0 & 21.0 & 49.0 & 26.0 & 53.0 & 18.0 & 29.3 \\
& SVD-LLM$^\dagger$      & 53.7 & 400 & 300 & 14.0 & 28.0 & 22.0 & 50.0 & 27.0 & 55.0 & 21.0 & 31.0  \\
& Dobi-SVD$^{*\dagger}$  & 46.2 & 200 & 200 & 15.0 & 31.0 & 20.0 & 52.0 & 28.0 & 54.0 & 22.0 & 31.7  \\
& SAES-SVD               & 22.0 & 117 & 94.0 & 16.0 & 33.0 & \textbf{25.0} & 52.0 & 30.0 & 54.0 & \textbf{23.0} & 33.3  \\
& \textbf{Ours (L3)}     & \textbf{11.4} & \textbf{62.8} & \textbf{57.2} & \textbf{17.4} & \textbf{41.2} & 22.4 & \textbf{55.9} & \textbf{32.2} & \textbf{57.9} & 22.1 & \textbf{35.6}  \\
\bottomrule
\end{tabular}
\caption{Comparison with SVD-family methods on LLaMA-7B.
Symbols follow~\cite{hu2026saes}: $\dagger$ denotes fine-tuning in the source protocol; $*$ denotes mixed-rank allocation.
All methods use 256 WikiText-2 calibration samples.
Baseline numbers from~\cite{hu2026saes}, our rows evaluated with \texttt{lm-eval-harness} v0.4. Zero-shot accuracy may differ across harness versions. 
DipSVD is absent at $\rho{=}0.6$ because the original paper
reports only $\rho \le 0.5$.}
\label{tab:svd-comparison}
\end{table*}

\section{Experiments}
\label{sec:experiments}

We evaluate the three-level chain along five dimensions:
overall performance against SVD-family baselines (\S\ref{sec:overall}),
sensitivity to compression ratio (\S\ref{sec:ratio-sensitivity}),
portability across architectures (\S\ref{sec:cross-arch}),
ablation of key design choices (\S\ref{sec:ablations}),
and compute/deployment efficiency (\S\ref{sec:efficiency}).

\subsection{Setup}

\paragraph{Baselines.}
We compare against five SVD-family baselines:
ASVD~\cite{yuan2023asvd},
SVD-LLM~\cite{wang2025svd},
Dobi-SVD~\cite{wang2025dobi},
DipSVD~\cite{ding2025dipsvd}, and
SAES-SVD~\cite{hu2026saes}.
Baseline numbers are from~\cite{hu2026saes} under a unified
evaluation protocol (same calibration data and evaluation
harness). Our numbers use the same pipeline with identical
hyperparameters except for the L2/L3 stages.
SVD-LLM + Sequential LoRA~\cite{wang2025svd}, which uses 50K
external Alpaca samples, is compared
in \S\ref{sec:overall}.
LiLlama~\cite{sy2024lillama} trains on 13M tokens. We compare
under our budget in Appendix~\ref{app:lillama-details}.

\paragraph{Models and Datasets.}
We evaluate on LLaMA-7B~\cite{touvron2023llama} at $\rho \in \{0.2, 0.4, 0.6, 0.8\}$.
Cross-architecture results on Mistral-7B~\cite{jiang2023mistral7b},
OPT-6.7B~\cite{zhang2022opt}, LLaMA-2-7B, and
LLaMA-2-13B~\cite{touvron2023llama2} appear in \S\ref{sec:cross-arch}.
All methods use the same 256 WikiText-2~\cite{merity2016pointer}
calibration sequences ($2048$ tokens each) from the
\emph{training} split.
Perplexity is evaluated on the \emph{test} splits of
WikiText-2 and PTB, and the C4 validation set.

\paragraph{Evaluation.}
We report WikiText-2, PTB, and C4 perplexity, and seven zero-shot
downstream tasks (OpenBookQA, ARC-e, ARC-c, WinoGrande, HellaSwag,
PIQA, MathQA) via \texttt{lm-evaluation-harness} v0.4.

\paragraph{Implementation Details.}
Our method applies a uniform rank per weight matrix (\S\ref{sec:l1}). Methods marked $*$ in Table~\ref{tab:svd-comparison} use their original non-uniform rank allocation. 
Lookahead probes $P_\ell$ are trained with Adam (lr $10^{-3}$, 200 steps). L2 uses Adam (lr $10^{-4}$) with early stopping per block. L3 uses AdamW (lr $2{\times}10^{-5}$, warmup\,10 + cosine decay) with early stopping on the WikiText-2 validation split. Full hyperparameters are in Appendix~\ref{app:hyperparams}.

\subsection{Overall Performance}
\label{sec:overall}

\paragraph{Comparison with SVD-Family Methods.}
 L3 achieves the lowest WikiText-2 perplexity and highest zero-shot average at every tested ratio
(Table~\ref{tab:svd-comparison}), with gains widening at aggressive compression.
At $\rho{=}0.6$, L3 nearly halves SAES-SVD's perplexity using only 256 calibration sequences, without external data, whereas
SVD-LLM + Sequential LoRA requires 50K Alpaca
samples~\cite{wang2025svd} yet reaches only 15.0.
Zero-shot accuracy also improves on most tasks
(Table~\ref{tab:svd-comparison}), though all compressed states stay
well below the dense model, so we read zero-shot accuracy as a
retention diagnostic rather than as evidence of preserved capability
(Appendix~\ref{app:robustness}).

\paragraph{Comparison with LiLlama.}
LiLlama~\cite{sy2024lillama} is the closest prior block-level SVD method, making it a natural test of whether our L2 gains stem from joint optimization itself or from the MSE + lookahead KL objective.
We reproduce its per-block distillation under our 256-sample budget
(Appendix~\ref{app:lillama-details}) and separate three distinct
properties.
On \emph{standalone quality}, our LiLlama-style reproduction is
stronger on every reported LLaMA-7B final metric, including
WikiText-2 perplexity ($10.5$ vs.\ $11.4$) and zero-shot average
($43.9\%$ vs.\ $35.6\%$; Table~\ref{tab:lillama}); we claim no
uniform quality advantage over it.
On \emph{offline cost}, our retained logs lack matched hardware
metadata, so a compute-matched comparison is unavailable and we make
no speedup claim.
On \emph{L3 composability}, under our matched $2{\times}10^{-5}$
refinement recipe on Mistral-7B, WikiText-2 validation selects
step~0 and the re-evaluated perplexity is $29.04/738.20/164.46$ on
WikiText-2/PTB/C4: a favourable local-recovery solution need not be
an L3-friendly warm start.
LiLlama's own budget is far larger---13M Slim-Orca tokens at 20\%
reduction, and 191M tokens for its reported 40\%-compressed Mistral
recovery---so we read this as warm-start and schedule sensitivity
under our 60\%-removal, 256-sequence protocol, not as a universal or
unfixable defect; spectral renormalization remains untested
(Figure~\ref{fig:spectral-sigma1}).

\begin{figure}[t]
\centering
\includegraphics[width=\columnwidth]{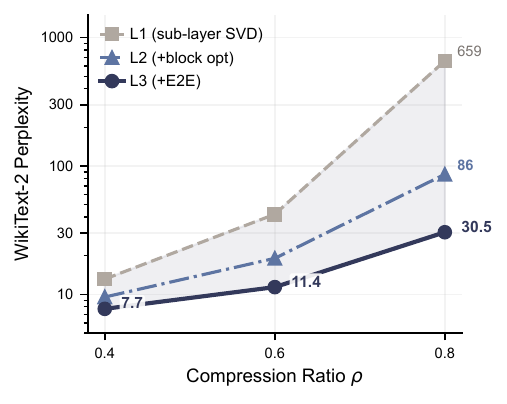}
\caption{WikiText-2 perplexity (log scale) vs.\ compression ratio on LLaMA-7B.
  The shaded region shows the hierarchy's cumulative gain (L1$\to$L3).
  L1 captures most of the gain at $\rho{=}0.2$, but all three levels are needed at $\rho{=}0.8$.}
\label{fig:ratio-sensitivity}
\end{figure}

\subsection{Ratio Sensitivity}
\label{sec:ratio-sensitivity}

The hierarchy's benefit grows with compression aggressiveness
(Figure~\ref{fig:ratio-sensitivity};
Appendix~\ref{app:ratios-table}).
L1 alone suffices at $\rho{=}0.2$, where L2/L3 add little. The gap widens at $\rho{=}0.6$ (L2 reduces L1's perplexity by ${\sim}2\times$. L3 closes half the remaining gap) and becomes essential at $\rho{=}0.8$, where the three levels
together yield a ${\sim}22\times$ cumulative reduction. In practice, L1+L2 captures most of the gain. L3 is most worthwhile when OOD robustness matters or compression exceeds 50\%.

\subsection{Cross-Architecture Portability}
\label{sec:cross-arch}

The chain improves all five models at $\rho{=}0.6$, though the per-level contributions are architecture-dependent (Table~\ref{tab:cross-arch}; Figure~\ref{fig:cross-arch}). On LLaMA-7B, L2 provides most of the OOD repair; L3 reduces perplexity further across all three test sets.
On Mistral-7B, L2 halves WikiText-2 perplexity, but L3 provides the larger downstream-accuracy gain. OPT-6.7B shows the largest collapse-and-recovery pattern, with L3 reaching the best downstream accuracy in the table.
LLaMA-2-13B extends the trend beyond 7B. LLaMA-2-7B is the outlier: WikiText-2 and C4 improve, but PTB remains high, partly reflecting the dense model's own domain sensitivity, though the large amplification warrants further investigation.
These rows were produced with architecture-specific configurations
rather than one common protocol, so we report the per-model outcomes
descriptively and do not attribute the differences among them to
architecture alone; separating those factors would require matched
interventions on grouped-query attention, sliding-window attention,
residual scaling, or other architectural choices.

\begin{table}[t]
\centering
\small
\setlength{\tabcolsep}{3pt}
\begin{tabular}{@{}llrrrr@{}}
\toprule
Model & Stage & Wiki$\downarrow$ & PTB$\downarrow$ & C4$\downarrow$ & Avg.\,$\uparrow$ \\
\midrule
\multirow{3}{*}{LLaMA-7B}
  & L1  & 42.1           & 182   & 201   & 33.2 \\
  & L2  & 19.1           & 95.8  & 91.8  & 35.3 \\
  & \textbf{L3}  & \textbf{11.4}  & \textbf{62.8} & \textbf{57.2} & \textbf{37.8} \\
\midrule
\multirow{3}{*}{Mistral-7B}
  & L1  & 71.8           & 770   & 439   & 31.0 \\
  & L2  & 34.0           & 371   & 182   & 32.7 \\
  & \textbf{L3}  & \textbf{14.2}  & \textbf{219}  & \textbf{85.5} & \textbf{36.1} \\
\midrule
\multirow{3}{*}{LLaMA-2-7B}
  & L1  & 81.5           & 1757  & 567   & 31.8 \\
  & L2  & 31.8           & 1438  & 169   & 33.9 \\
  & \textbf{L3}  & \textbf{18.1}  & \textbf{1235} & \textbf{87.1} & \textbf{34.3} \\
\midrule
\multirow{3}{*}{OPT-6.7B}
  & L1  & 42.9           & 1889  & 282   & 33.9 \\
  & L2  & 26.0           & 79.6  & 96.0  & 36.8 \\
  & \textbf{L3}  & \textbf{18.3}  & \textbf{52.9} & \textbf{68.3} & \textbf{39.7} \\
\midrule
\multirow{3}{*}{LLaMA-2-13B}
  & L1  & 42.3           & 1042  & 277   & 33.0 \\
  & L2  & 21.8           & 615   & 119   & 34.7 \\
  & \textbf{L3}  & \textbf{11.2}  & \textbf{513} & \textbf{62.4} & \textbf{37.4} \\
\bottomrule
\end{tabular}
\caption{Cross-architecture L1$\to$L2$\to$L3 at $\rho{=}0.6$.
Perplexity on WikiText-2, PTB, and C4; Avg.\ is zero-shot accuracy
over ARC-e, HellaSwag, MathQA, OBQA, PIQA, WinoGrande.
WikiText-2 trend visualized in Figure~\ref{fig:cross-arch}.}
\label{tab:cross-arch}
\end{table}

\subsection{Compute and Deployment Efficiency}
\label{sec:efficiency}

The three-level chain spends offline compute to gain compression
fidelity, and the factored model reduces parameter memory at
inference (Table~\ref{tab:efficiency}).
Our retained run logs do not record hardware metadata, so we report
correctly scoped per-stage timers rather than a
hardware-attributed total: averaged over three runs, L1 takes
$17.3$\,min, L2 $13.76$\,h, and L3 $42.4$\,min, with the sequential
per-block loop accounting for $13.59$\,h ($98.8\%$) of L2 and probe
fitting roughly $6$\,min. The block loop, not probe fitting, therefore
dominates the block-level stage.
These timers exclude loading, serialization, and final perplexity
evaluation, and the implementation is unoptimized, so they do not
establish a hardware-matched break-even point against training-free
baselines; latency amortization remains unquantified.
At inference, the factored representation reduces parameter count
by ${\sim}58\%$ and peak GPU memory by $49$--$56\%$ across architectures.
Prefill latency is unchanged; batch-1 decode is
$6$--$14\%$ slower (Table~\ref{tab:efficiency}) because each projection
executes two unfused GEMMs, which adds kernel-launch and
memory-access overhead. The slowdown is thus specific to this eager,
unfused factorized implementation rather than inherent to low-rank
arithmetic; kernel fusion, CUDA graphs, and compilation are plausible
remedies we have not validated. The demonstrated deployment-side
benefit is lower parameter memory and peak VRAM, and we make no
general deployment-time claim.
The factored matrices $A_j, B_j$ can also be quantized to INT8 or
INT4 for additional savings, as low-rank projection and
quantization~\cite{frantar2022gptq} occupy orthogonal points on
the compression Pareto frontier.

\begin{table}[t]
\centering
\small
\setlength{\tabcolsep}{3pt}
\begin{tabular}{@{}llrrrr@{}}
\toprule
Model & & Par.\,(M)$\downarrow$ & Mem\,(GB) & Pref.\,(ms) & Gen$\uparrow$ \\
\midrule
\multirow{2}{*}{LLaMA-7B}
  & Base & 6738 & 14.6\,G & 286 & 43.2 \\
  & L3   & 2853 & 7.4\,G  & 284 & 39.3 \\
\midrule
\multirow{2}{*}{Mistral-7B}
  & Base & 7248 & 14.9\,G & 287 & 41.8 \\
  & L3   & 3060 & 7.1\,G  & 275 & 37.7 \\
\midrule
\multirow{2}{*}{OPT-6.7B}
  & Base & 6659 & 14.5\,G & 254 & 60.8 \\
  & L3   & 2792 & 6.3\,G  & 256 & 52.1 \\
\bottomrule
\end{tabular}
\caption{Deployment efficiency at $\rho{=}0.6$ (fp16, bs=1,
seq=2048). Gen = decode throughput (tok/s, 64 tokens).
Relative decode throughput in the same eager benchmark is
$0.91\times$ (LLaMA-7B), $0.90\times$ (Mistral-7B), $0.94\times$
(LLaMA-2-7B) and $0.86\times$ (OPT-6.7B); no comparable
LLaMA-2-13B measurement is available.}
\label{tab:efficiency}
\end{table}

\subsection{Ablation Study}
\label{sec:ablations}

\paragraph{Target Choice and Lookahead.}
The lookahead probe is the single most impactful design choice in
L2, amplifying perplexity reduction by $2{-}3\times$ over target choice
alone (Table~\ref{tab:target-ablation}). Without LA, both targets reduce Mistral-7B WikiText-2 by only
20--23\% relative to L1; with LA the reduction reaches 53--59\%
(Table~\ref{tab:target-ablation}). On LLaMA-7B, LA further improves already-helpful no-LA variants.
We adopt CF+LA as the default: SI+LA wins on Mistral-7B, but CF+LA provides the best single configuration across both architectures (Table~\ref{tab:target-ablation}). We attribute the architecture gap to the fact that block MSE
weights all hidden directions equally, while the downstream loss
amplifies errors in directions with small hidden-state norms. The lookahead probe's vocabulary-space KL signal addresses this mismatch.

\begin{table}[t]
\centering
\begin{subtable}{\columnwidth}
\centering
\small
\setlength{\tabcolsep}{3pt}
\begin{tabular}{@{}llcccc@{}}
\toprule
Model & Config & Target & LA & Wiki$\downarrow$ & PTB$\downarrow$ \\
\midrule
\multirow{5}{*}{Mistral-7B}
  & L1 baseline  & ---  & ---  & 71.8  & 770  \\
  & L2 (SI)      & SI   & no   & 57.5  & 631  \\
  & L2 (CF)      & CF   & no   & 55.6  & 649  \\
  & \textbf{L2 (SI+LA)} & SI & yes & \textbf{29.5} & \textbf{318} \\
  & L2 (CF+LA)   & CF   & yes  & 34.0  & 374  \\
\midrule
\multirow{5}{*}{LLaMA-7B}
  & L1 baseline  & ---  & ---  & 42.1  & 182  \\
  & L2 (SI)      & SI   & no   & 38.0  & 253  \\
  & L2 (CF)      & CF   & no   & 35.4  & 275  \\
  & L2 (SI+LA)   & SI   & yes  & 28.3  & 165  \\
  & \textbf{L2 (CF+LA)} & CF & yes & \textbf{24.8} & \textbf{146} \\
\bottomrule
\end{tabular}
\caption{Target and lookahead ablation ($\rho{=}0.6$, 20 Adam steps/block).
  SI = same-input: target is $f_\ell(h_\ell^{\text{comp}})$;
  CF = cross-frontier: target is $h_{\ell+1}^{\text{orig}}$;
  LA = lookahead probe (\S\ref{sec:method}).}
\label{tab:target-ablation}
\end{subtable}
\vspace{6pt}
\begin{subtable}{\columnwidth}
\centering
\small
\setlength{\tabcolsep}{3pt}
\begin{tabular}{@{}lccccc@{}}
\toprule
Path & LR & Step & Wiki$\downarrow$ & PTB$\downarrow$ & C4$\downarrow$ \\
\midrule
L1$\to$L2$\to$L3 & 5e-6 & 100 & 11.95 & 64.74  & 52.21 \\
\textbf{L1$\to$L2$\to$L3} & \textbf{2e-5} & \textbf{40} & \textbf{11.37} & \textbf{62.67}  & \textbf{57.24} \\
\midrule
L1$\to$L3$'$ & 5e-6 & 90  & 11.91 & 107.79 & 58.51 \\
L1$\to$L3$'$ & 2e-5 & 100 & 11.90 & 86.87  & 60.24 \\
\bottomrule
\end{tabular}
\caption{Effect of L2 under a fixed 256-sample budget.
L3$'$: LM-loss applied directly to L1 factors.
These rows come from the submitted learning-rate sweep;
Table~\ref{tab:pool-sweep} reports the matched fixed-budget
comparison used for the data-versus-compute analysis, and
Table~\ref{tab:seeds} the corresponding multi-seed check.}
\label{tab:skip_level2}
\end{subtable}
\caption{Ablation studies ($\rho{=}0.6$).}
\label{tab:ablations}
\end{table}

\paragraph{Block-Level Optimization.}
L2 is not redundant with L3: skipping it preserves in-distribution
perplexity, but degrades OOD transfer by 24 points on PTB (Table~\ref{tab:skip_level2}). The full chain early-stops sooner with better OOD metrics, while skip-L2 trains longer yet produces worse OOD perplexity (Figure~\ref{fig:skip-l2-dynamics}): additional LM-loss training did not recover the regularization that L2's hidden-state matching provides under the tested learning-rate schedules.
This direction is not specific to one calibration draw: across three
end-to-end pipeline seeds, in which each seed resamples L1
calibration sampling and L2/L3 randomness, the full chain attains
lower WikiText-2, PTB and C4 perplexity than skip-L2 for every seed
(Table~\ref{tab:seeds}). Three seeds establish a consistent
direction with variable effect size, not low variance or
arbitrary-seed robustness.

\paragraph{Calibration Budget.}

\begin{figure}[t]
\centering
\includegraphics[width=\columnwidth]{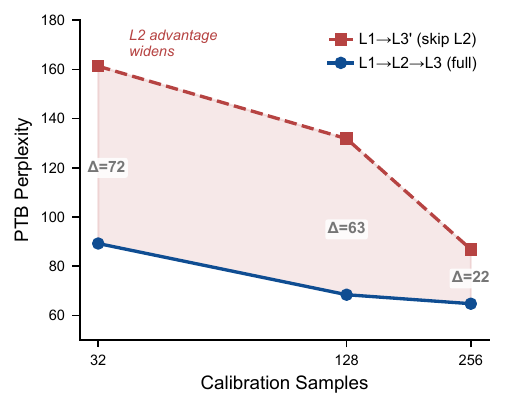}
\caption{PTB test perplexity vs.\ calibration budget (LLaMA-7B, $\rho{=}0.6$). L2's OOD advantage grows from $\Delta{=}22$ at 256 samples to $\Delta{=}72$ at 32 samples.}
\label{fig:data-efficiency}
\end{figure}

Block-level optimization caches full hidden states, so practical deployments may be limited to small calibration sets. L2's OOD advantage grows as calibration data shrinks (Figure~\ref{fig:data-efficiency}): the PTB gap between the full chain and skip-L2 more than triples from 256 to 32 samples.
L2's block-level regularization is most valuable under data
scarcity (Appendix~\ref{app:lowdata}).

\paragraph{Can More Data Replace L2?}
Because L2 spends offline compute rather than data, the natural
alternative is to skip it and enlarge the calibration pool instead.
On the seed-42 path this is effective
(Table~\ref{tab:pool-sweep}): skip-L2 at 1024 sequences reaches
lower WikiText-2 and PTB perplexity than the full chain at 256,
while remaining $1.18$ PPL worse on C4.
An exposure control rules out the longer schedule as the
explanation---running skip-L2 at pool 256 for the same 160 optimizer
steps, 640 draws and $1.311$M tokens as the pool-1024 run leaves the
WikiText-2 selection rule choosing the elementwise-identical
step-45 factors.
We therefore read L2 as trading offline compute for calibration
data rather than as uniformly necessary: it helps under a strict
256-sequence budget, while a larger pool makes skip-L2 a competitive
cross-budget alternative. The full chain at pool 1024 remains
untested.
Expressed as additive next-token negative log-likelihood, the mean
full-versus-skip gain at pool 256 is $0.070/0.335/0.172$
nats/token on WikiText-2/PTB/C4, against $0.112/0.332/0.184$ for
moving skip-L2 from pool 256 to 1024---comparable in magnitude,
which is what makes the two routes substitutable at this budget.

\begin{table}[t]
\centering
\small
\setlength{\tabcolsep}{3pt}
\begin{tabular}{@{}lcccc@{}}
\toprule
Path & Pool & Wiki$\downarrow$ & PTB$\downarrow$ & C4$\downarrow$ \\
\midrule
Skip-L2    & 256  & 11.97 & 91.44 & 59.71 \\
Skip-L2    & 512  & 11.30 & 67.15 & 56.77 \\
Skip-L2    & 1024 & \textbf{10.71} & \textbf{65.61} & 49.70 \\
\midrule
Full chain & 256  & 11.41 & 67.64 & \textbf{48.53} \\
\bottomrule
\end{tabular}
\caption{Trading calibration data for the block-level stage
(LLaMA-7B, $\rho{=}0.6$, seed-42 path).
Enlarging the pool lets skip-L2 match or beat the full chain on
WikiText-2 and PTB, but not on C4.
Within each pair the arms share L1 factors, rank allocation, the L3
seed and schedule, and the WikiText-2 selection rule; the designed
difference is the presence of L2.}
\label{tab:pool-sweep}
\end{table}

\paragraph{Level Contributions.}
  Each level addresses a distinct error scale: L1 provides per-matrix initialization, L2 captures cross-sub-layer interactions within each block, and L3 corrects cross-block accumulation.

\section{Discussion}
\label{sec:discussion}

\paragraph{Compression Error at Multiple Interaction Scales.}
Low-rank compression error arises at distinct structural
scales, and block-internal coupling is the largest single source
of recoverable degradation.
The L1$\to$L2 transition accounts for more than half of the
total perplexity reduction on LLaMA-7B, while L2$\to$L3 closes
much of the remaining gap (\S\ref{sec:overall}).
The balance is architecture-dependent (on Mistral-7B, L3
provides the larger gain; Table~\ref{tab:cross-arch}), but L2
helps on all five models.

\paragraph{L2 as OOD Regularizer.}
Block-level optimization also serves a role that perplexity
alone does not reveal: it preserves out-of-distribution transfer.
Block-level MSE forces the compressed block to reproduce the
hidden-state distribution of the original, including directions
that the LM loss on calibration tokens alone does not reward,
analogous to ``dark knowledge'' in soft-target
distillation~\cite{hinton2015distilling}.
The skip-L2 ablation (\S\ref{sec:ablations}) confirms this:
in-distribution perplexity matches, yet PTB degrades by 24
points, and the penalty triples under data scarcity
(Figure~\ref{fig:data-efficiency}), consistent with the
spectral-norm amplification observed when L2 uses a
scale-invariant loss (Appendix~\ref{app:lillama-details}).

\paragraph{When to Use Each Level.}
Together, these observations suggest a practical guideline.
At mild compression ($\rho \le 0.2$), L1 alone captures most of the
gain, though L3 still improves WikiText-2 perplexity.
At moderate compression ($\rho{=}0.4$--$0.6$), L2 captures most
of the robustness gain. L3 is worthwhile when OOD behavior matters or computation permits. At aggressive compression ($\rho \ge 0.8$) the
full chain is needed; under severe calibration scarcity the
block-level stage still protects PTB but no longer dominates on
WikiText-2 or C4 (Table~\ref{tab:lowdata}).
These guidelines derive from LLaMA-7B and may shift for other architectures. 

\section{Conclusion}
\label{sec:conclusion}

We presented a three-level chain that widens SVD-based compression
from individual matrices to Transformer blocks to the full model,
using only 256 calibration sequences.
On LLaMA-7B at 60\% compression, the chain reduces WikiText-2 perplexity from 42.1 to 11.4, below SVD-LLM + 50K-sample LoRA~(15.0), with gains across five architectures whose runs use architecture-specific configurations rather than one common protocol.
The block-level stage matters for OOD robustness under this budget: skipping it costs 24 points on PTB, and the penalty grows as calibration data shrink; the direction repeats across three end-to-end pipeline seeds. Given a larger calibration pool, however, skipping it becomes competitive, so the stage is best read as trading offline compute for calibration data rather than as indispensable.
The gap closes not by a better decomposition but by optimizing at the right scope. We restrict these conclusions to perplexity and compression fidelity.

\section*{Limitations}
\label{sec:limitations}

Our evaluation covers five models at the 7B--13B scale.
We have not tested 70B models, instruction-following,
safety, or long-context settings.
Block-level optimization stores full hidden states per sample
(one hidden-state matrix per block boundary in fp16), limiting calibration to 256 samples. 
The lookahead probe provides only one-block-ahead sensitivity. Multi-block cascades are addressed by L3 rather than L2. L2 matches block outputs pointwise in the dense activation space; we did not test alternatives that supervise hidden states through transition geometry~\cite{li2026phf} or in a sparse autoencoder basis~\cite{zhang2026saefd}, proposed for self-distillation and continual learning respectively. Our calibration sequences are also drawn at random rather than selected; coverage-based selection criteria from the reinforcement-learning data setting~\cite{li2026irds} may change how much calibration data the block-level stage can be traded for. The factored representation slows decode throughput by 9--14\% (Table~\ref{tab:efficiency}) due to two sequential  matmuls per sub-layer; fused AB-kernels would close this gap but are not yet
implemented.
Downstream accuracy does not uniformly improve despite perplexity
gains (Table~\ref{tab:svd-comparison}), suggesting that the calibration distribution determines which capabilities survive
compression. 
The full-versus-skip comparison was repeated across three end-to-end
pipeline seeds (Appendix~\ref{app:robustness}), which shows a
consistent direction with variable effect size; this does not
establish low variance or arbitrary-seed robustness, and all other
reported results use seed 42.
Our conclusions are also sensitive to the calibration budget: the
block-level stage's advantage is established under the 256-sequence
regime, and enlarging the pool makes skipping it competitive
(Table~\ref{tab:pool-sweep}).
We claim only perplexity and compression fidelity. We do not claim
preservation of broad capability, reasoning, instruction following,
safety, or open-ended generation: at $60\%$ removal every compressed
state scores $0\%$ strict exact match on GSM8K and produces
degenerate, highly repetitive summaries on our held-out SAMSum
subset (Appendix~\ref{app:robustness}).
We also report no hardware-matched compute comparison or
break-even point, and the measured decode slowdown reflects our
eager, unfused factorized implementation rather than low-rank
arithmetic itself.
Combining low-rank factorization with post-training quantization
(e.g., GPTQ~\cite{frantar2022gptq}) is a natural next step, but interaction effects under aggressive settings are untested. Compression can unevenly degrade safety-related behaviors (e.g., refusal of harmful prompts), so deploying compressed models requires dedicated safety evaluation.

\section*{Acknowledgments}
This work was partially supported by the National Key R\&D Program of China (Grant No.~2024YFB4504004), the National Natural Science Foundation of China (Grant. No.~92576114, 12404568), the Guangdong Provincial Quantum Science Strategic Initiative (Grant No.~GDZX2403008, GDZX2503001), and the Modern Matter Lab (MML) at HKUST(GZ).

\appendix

\FloatBarrier
\section{LiLlama Comparison and Spectral Analysis}
\label{app:lillama-details}

\begin{table}[h!]
\centering
\small
\setlength{\tabcolsep}{3pt}
\resizebox{\columnwidth}{!}{%
\begin{tabular}{@{}llrrrrrc@{}}
\toprule
Model & Method & Wiki$\downarrow$ & PTB$\downarrow$ & C4$\downarrow$ & Avg.\,$\uparrow$ & Time \\
\midrule
\multirow{4}{*}{LLaMA-7B}
  & LiLlama$^*$ L2            & 12.4  & 56.2  & 39.6  & 42.2 & $\sim$48h \\
  & LiLlama$^*$ L2$\to$L3     & 10.5  & 58.4  & 48.0  & 43.9 & +$\sim$1h \\
  & Ours L2 (CF+LA)            & 19.1  & 95.8  & 91.8  & 34.6 & $\sim$12h \\
& Ours L3  & 11.4 & 62.8 & 57.2 & 35.6 & +$\sim$1h \\
\midrule
\multirow{4}{*}{Mistral-7B}
  & LiLlama$^*$ L2            & 28.7  & 743   & 161   & 33.9 & $\sim$48h \\
  & LiLlama$^*$ L2$\to$L3     & 29.0$^\dagger$  & 738   & 165   & 33.9 & +$\sim$1h \\
  & Ours L2 (CF+LA)            & 34.0  & 371   & 182   & 32.7 & $\sim$14h \\
  & \textbf{Ours L3}           & \textbf{14.2} & \textbf{219} & \textbf{85.5} & \textbf{36.7} & +$\sim$1h \\
\bottomrule
\multicolumn{7}{@{}l}{\scriptsize $^\dagger$Diverged: best checkpoint is step~0 (no improvement over L2).}
\end{tabular}}
\caption{Comparison with LiLlama-style per-block distillation at $\rho{=}0.6$,
both using 256 calibration samples.
LiLlama$^*$ denotes our reproduction of LiLlama's teacher+student (T+S) loss
(LiLlama's original paper uses 13M tokens).
Times are archived wall-clock readings whose logs do not record
hardware metadata; they are indicative only and do not support a
compute-matched comparison.
Avg.\ uses the 7-task set from Table~\ref{tab:svd-comparison}.}
\label{tab:lillama}
\end{table}

\paragraph{Re-Extraction Control.}
LiLlama's L2$\to$L3 path requires SVD re-extraction (LiLlama saves
merged dense weights), whereas our pipeline restores saved CompressedLinear factors.
To rule out re-extraction as the cause of LiLlama's divergence, we apply the same merge$\to$re-extract$\to$F-stage recipe to our own
CF+LA model: WikiText-2 improves from $11.4$ (saved factors) to $11.37$
(re-extracted) on LLaMA-7B.
Re-extraction with QR$+$SVD rebalancing is thus benign or mildly
beneficial---LiLlama's divergence correlates with spectral amplification, not the re-extraction procedure.

\paragraph{OOD Transfer.}
On LLaMA-7B, LiLlama's L3 \emph{worsens} C4 perplexity
($+21\%$) despite improving WikiText-2, while our L3 improves
both (Table~\ref{tab:lillama}).
We attribute this to \emph{headroom}: LiLlama's L2 already
approaches the dense baseline on WikiText-2, leaving little room
for WikiText-2-calibrated L3 to improve without overfitting to the
calibration domain.
Our weaker L2 has ample headroom, so L3 gains generalize to
held-out corpora.

\begin{figure*}[t]
\centering
\includegraphics[width=\textwidth]{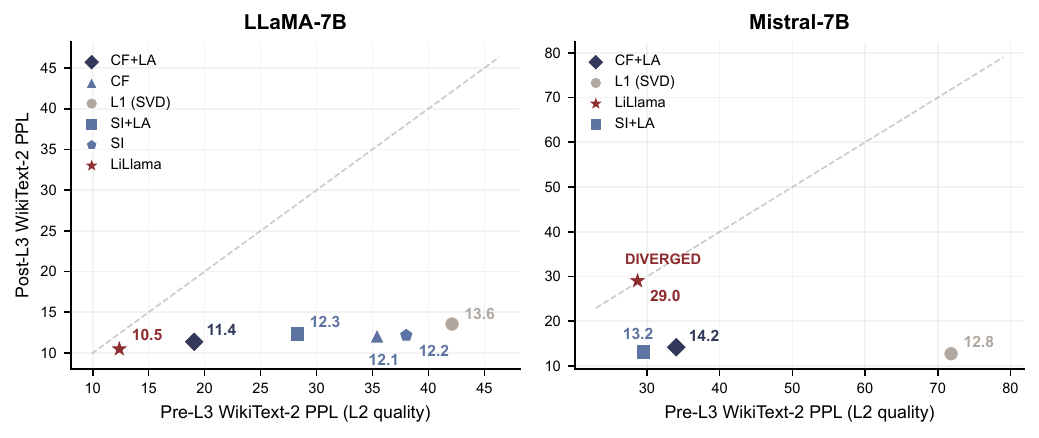}
\caption{Pre-L3 vs.\ post-L3 WikiText-2 test perplexity
($\rho{=}0.6$, 256 calibration samples, best val checkpoint).
Left: LLaMA-7B---better L2 warm-start yields better L3 (monotonic).
Right: Mistral-7B---LiLlama achieves the best L2 but diverges at L3,
coinciding with the spectral amplification of
Figure~\ref{fig:spectral-sigma1}.}
\label{fig:scatter-warmstart}
\end{figure*}

\paragraph{Warm-Start Quality.}
Figure~\ref{fig:scatter-warmstart} visualizes pre-L3 vs.\ post-L3
WikiText-2 perplexity across warm-start constructors (L1, SI, CF,
CF+LA, LiLlama).
On LLaMA-7B the relationship is monotonic (better L2 $\Rightarrow$
better L3). On Mistral-7B it breaks---LiLlama achieves the best L2
but diverges at L3.

\begin{figure*}[t]
\centering
\includegraphics[width=\textwidth]{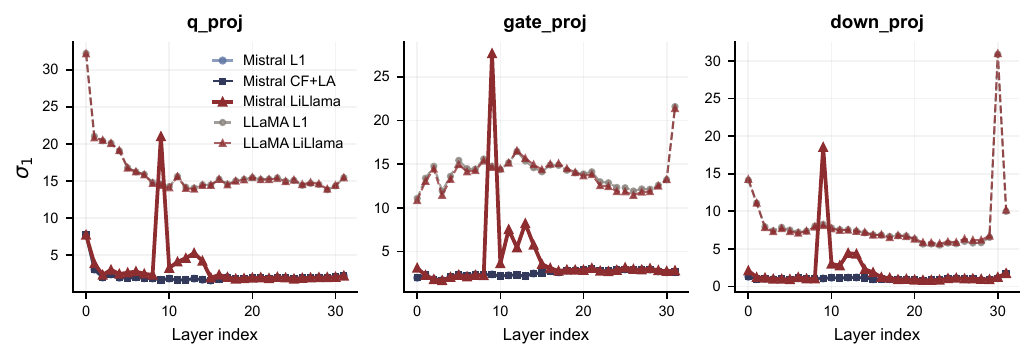}
\caption{Leading singular value $\sigma_1$ of compressed weights
by layer ($\rho{=}0.6$). LiLlama (red, solid) amplifies Mistral-7B
MLP $\sigma_1$ by 40--95\% vs.\ L1 baseline (light blue). Our CF+LA (dark blue) preserves the baseline spectral structure. LLaMA-7B (dashed) shows milder amplification.}
\label{fig:spectral-sigma1}
\end{figure*}

\paragraph{Spectral Stability.}
Figure~\ref{fig:spectral-sigma1} shows the leading singular value
$\sigma_1$ of compressed weight matrices across all 32 layers for three
sub-layer types.
Cosine-based per-block distillation (the T+S loss used by
LiLlama~\cite{sy2024lillama}) is scale-invariant, allowing $\sigma_1$ to
grow unchecked during L2 optimization.
Under our 256-sample budget, Mistral-7B MLP sub-layers
(gate\_proj, down\_proj) exhibit 40--95\% $\sigma_1$ amplification
relative to the L1 baseline, correlating with gradient explosion
during subsequent L3 training.
This instability may be specific to the low-data regime.
LiLlama's original 13M-token budget likely provides sufficient
implicit regularization to prevent it.
Our MSE-based L2 implicitly regularizes spectral norms, keeping
$\sigma_1$ stable across architectures and data budgets.

\section{Full Results Tables}
\label{app:ratios-table}

Table~\ref{tab:ratios} reports perplexity and zero-shot accuracy for all four compression ratios on LLaMA-7B. Each level improves at every ratio. The gap widens at aggressive compression ($\rho{=}0.8$: L1 gives 659, L3 reaches 30.5).

\begin{table*}[t]
\centering
\footnotesize
\setlength{\tabcolsep}{3pt}
\begin{tabular}{@{} cl rrr cccccccc c @{}}
\toprule
& & \multicolumn{3}{c}{Perplexity $\downarrow$} & \multicolumn{9}{c}{Zero-Shot Accuracy (\%) $\uparrow$} \\
\cmidrule(lr){3-5} \cmidrule(lr){6-14}
$\rho$ & Stage & Wiki & PTB & C4 & ARC\textsubscript{c} & ARC\textsubscript{e} & BoolQ & HellaS. & Openb. & PIQA & WinoG. & TruthQA & MathQA \\
\midrule
--- & \emph{Uncompressed} & \emph{5.68} & \emph{8.35} & \emph{7.34} & \emph{41.8} & \emph{75.3} & \emph{75.0} & \emph{57.0} & \emph{34.6} & \emph{78.7} & \emph{69.9} & \emph{34.1} & \emph{27.1} \\
\midrule
\multirow{3}{*}{0.2}
& L1 & 7.86 & 15.4 & 15.9 & 31.6 & 62.7 & \textbf{64.4} & 43.2 & 26.6 & 68.8 & \textbf{67.2} & \textbf{37.7} & 23.4 \\
& L2 & 7.00 & \textbf{13.5} & \textbf{13.3} & 34.7 & \textbf{69.9} & 63.3 & 46.5 & 28.8 & 71.2 & 65.6 & \textbf{37.7} & 25.7 \\
& \textbf{L3} & \textbf{6.47} & 14.1 & 13.5 & \textbf{35.0} & \textbf{69.9} & 63.3 & \textbf{47.6} & \textbf{29.8} & \textbf{71.7} & 66.0 & 37.4 & \textbf{26.5} \\
\midrule
\multirow{3}{*}{0.4}
& L1 & 13.16 & 52.5 & 50.0 & 27.0 & 45.9 & 43.0 & 33.0 & 20.2 & 60.9 & 57.1 & \textbf{43.3} & 22.1 \\
& L2 & 9.54 & \textbf{31.1} & 27.8 & 27.6 & 55.9 & \textbf{63.2} & 37.6 & 22.6 & 63.9 & \textbf{62.1} & 42.8 & 23.5 \\
& \textbf{L3} & \textbf{7.72} & 31.3 & \textbf{25.7} & \textbf{29.9} & \textbf{59.8} & 63.0 & \textbf{40.4} & \textbf{24.2} & \textbf{65.6} & 62.0 & 42.4 & \textbf{24.4} \\
\midrule
\multirow{3}{*}{0.6}
& L1 & 42.1 & 182 & 201 & 19.5 & 29.9 & 37.8 & 27.3 & 13.2 & 54.3 & 52.4 & 47.6 & 21.9 \\
& L2 & 19.1 & 95.8 & 91.8 & 20.2 & 34.6 & 37.8 & 29.0 & 14.6 & 56.4 & 54.8 & \textbf{49.0} & \textbf{22.1} \\
& \textbf{L3} & \textbf{11.4} & \textbf{62.8} & \textbf{57.2} & \textbf{22.4} & \textbf{41.2} & \textbf{54.8} & \textbf{32.2} & \textbf{17.4} & \textbf{57.9} & \textbf{55.9} & 48.3 & \textbf{22.1} \\
\midrule
\multirow{3}{*}{0.8}
& L1 & 659 & 3687 & 2638 & \textbf{20.5} & 26.2 & 37.8 & 26.0 & \textbf{13.2} & 52.3 & 48.7 & \textbf{50.2} & 20.6 \\
& L2 & 86.2 & 526 & 410 & 19.4 & \textbf{27.7} & 37.8 & 26.8 & 12.2 & 52.9 & 49.7 & 49.3 & 20.8 \\
& \textbf{L3} & \textbf{30.5} & \textbf{158} & \textbf{143} & 19.1 & \textbf{27.7} & \textbf{37.9} & \textbf{27.2} & 12.4 & \textbf{53.8} & \textbf{51.7} & 48.8 & \textbf{20.9} \\
\bottomrule
\end{tabular}
\caption{Three-level hierarchical compression on LLaMA-7B, WikiText-2 calibration, seed 42.
\textbf{L1}: sub-layer SVD;
\textbf{L2}: block-level joint optimization with 256-sample block I/O;
\textbf{L3}: end-to-end LM-loss fine-tune on top of L2.
Dense baseline from our lm-eval harness (adds BoolQ and TruthfulQA to Table~\ref{tab:svd-comparison}'s 7-task set).}
\label{tab:ratios}
\end{table*}

Table~\ref{tab:cross-arch-full} reports dense baselines alongside
our L3 per-task zero-shot accuracy for all five architectures.
Dense PTB/Wiki ratios range from $1.2\times$ (OPT-6.7B) to $7.2\times$ (LLaMA-2-13B), suggesting that high compressed-model  PTB partly reflects pre-existing domain sensitivity, though the amplification after compression remains a limitation.

\begin{table*}[t]
\centering
\footnotesize
\setlength{\tabcolsep}{4pt}
\begin{tabular}{@{}ll rrr cccccc r@{}}
\toprule
& & \multicolumn{3}{c}{PPL $\downarrow$} & \multicolumn{6}{c}{Zero-Shot Acc.\ (\%) $\uparrow$} & \\
\cmidrule(lr){3-5} \cmidrule(lr){6-11}
Model & & Wiki & PTB & C4 & ARC\textsubscript{e} & HellaS. & MathQA & OBQA & PIQA & WinoG. & Avg. \\
\midrule
\multirow{2}{*}{LLaMA-7B}
  & \emph{Dense} & \emph{5.68} & \emph{8.35} & \emph{7.34} & \emph{67.0} & \emph{56.0} & \emph{27.0} & \emph{28.0} & \emph{78.0} & \emph{67.0} & \emph{53.8} \\
  & \textbf{Ours} & \textbf{11.4} & \textbf{62.8} & \textbf{57.2} & 41.2 & 32.2 & 22.1 & 17.4 & 57.9 & 55.9 & \textbf{37.8} \\
\midrule
\multirow{2}{*}{Mistral-7B}
  & \emph{Dense} & \emph{5.31} & \emph{30.1} & \emph{8.14} & \emph{80.2} & \emph{61.1} & \emph{37.2} & \emph{35.4} & \emph{80.0} & \emph{74.1} & \emph{61.3} \\
  & \textbf{Ours} & \textbf{14.2} & \textbf{219} & \textbf{85.5} & 36.5 & 30.3 & 22.2 & 16.8 & 56.5 & 54.5 & \textbf{36.1} \\
\midrule
\multirow{2}{*}{LLaMA-2-7B}
  & \emph{Dense} & \emph{5.47} & \emph{25.2} & \emph{7.13} & \emph{75.4} & \emph{57.1} & \emph{28.3} & \emph{33.4} & \emph{78.1} & \emph{69.3} & \emph{56.9} \\
  & \textbf{Ours} & \textbf{18.1} & \textbf{1235} & \textbf{87.1} & 32.0 & 29.0 & 21.9 & 15.6 & 54.5 & 52.6 & \textbf{34.3} \\
\midrule
\multirow{2}{*}{OPT-6.7B}
  & \emph{Dense} & \emph{10.9} & \emph{12.8} & \emph{12.3} & \emph{66.1} & \emph{50.3} & \emph{24.2} & \emph{26.4} & \emph{76.5} & \emph{65.5} & \emph{51.5} \\
  & \textbf{Ours} & \textbf{18.3} & \textbf{52.9} & \textbf{68.3} & 45.6 & 33.1 & 22.0 & 20.2 & 60.7 & 56.4 & \textbf{39.7} \\
\midrule
\multirow{2}{*}{LLaMA-2-13B}
  & \emph{Dense} & \emph{4.88} & \emph{35.3} & \emph{6.62} & \emph{78.9} & \emph{60.2} & \emph{31.1} & \emph{34.6} & \emph{79.4} & \emph{72.6} & \emph{59.5} \\
  & \textbf{Ours} & \textbf{11.2} & \textbf{513} & \textbf{62.4} & 38.6 & 32.3 & 22.0 & 18.2 & 58.1 & 55.4 & \textbf{37.4} \\
\bottomrule
\end{tabular}
\caption{Dense baselines and per-task zero-shot accuracy at $\rho{=}0.6$.
Extends Table~\ref{tab:cross-arch}.}
\label{tab:cross-arch-full}
\end{table*}

\section{Low-Data Ablation and Hyperparameters}
\label{app:lowdata}

Table~\ref{tab:lowdata} shows how the OOD advantage of the full chain grows as calibration data shrinks.
At 32 samples, skipping L2 degrades PTB by 72 points (161 vs.\ 89) compared with 22 points at 256 samples, confirming that L2's block-level regularization is most valuable under data scarcity.
Table~\ref{tab:hyperparams} lists the full hyperparameters for L2 and L3.

\begin{table*}[t]
\centering
\begin{minipage}[t]{0.48\textwidth}
\centering
\small
\setlength{\tabcolsep}{3pt}
\begin{tabular}{@{}clccc@{}}
\toprule
Samples & Path & Wiki$\downarrow$ & PTB$\downarrow$ & C4$\downarrow$ \\
\midrule
\multirow{3}{*}{256}
& L1$\to$L3$'$ \scriptsize{(skip L2)}  & 11.90 & 86.87  & 60.24  \\
& L1$\to$L2 only                 & 21.59 & 110.09 & 104.97 \\
& L1$\to$L2$\to$L3 \scriptsize{(full)}  & 11.95 & 64.74  & 52.21  \\
\midrule
\multirow{3}{*}{128}
& L1$\to$L3$'$ \scriptsize{(skip L2)}  & 12.72 & 131.83 & 62.19  \\
& L1$\to$L2 only                 & 21.71 & 112.09 & 107.38 \\
& L1$\to$L2$\to$L3 \scriptsize{(full)}  & 12.57 & 68.40  & 56.58  \\
\midrule
\multirow{3}{*}{32}
& L1$\to$L3$'$ \scriptsize{(skip L2)}  & 14.87 & 161.30 & 73.18  \\
& L1$\to$L2 only                 & 26.82 & 141.83 & 132.90 \\
& L1$\to$L2$\to$L3 \scriptsize{(full)}  & 16.54 & 89.22  & 87.64  \\
\bottomrule
\end{tabular}
\caption{Low-data ablation ($\rho{=}0.6$, Wiki/PTB/C4 perplexity).
Full chain uses lr $5{\times}10^{-6}$; L3$'$ (skip L2) uses lr $2{\times}10^{-5}$
(each path's best LR).}
\label{tab:lowdata}
\end{minipage}%
\hfill
\begin{minipage}[t]{0.48\textwidth}
\centering
\small
\setlength{\tabcolsep}{2pt}
\begin{tabular}{@{}lp{2.2cm}p{2.8cm}@{}}
\toprule
Parameter & L2 (per block) & L3 (end-to-end) \\
\midrule
Optimizer       & Adam          & AdamW \\
Learning rate   & $10^{-4}$     & $2{\times}10^{-5}$ \\
LR schedule     & constant      & warmup\,10 + cosine \\
Weight decay    & 0             & 0 \\
Grad clip       & norm 1.0      & norm 1.0 \\
Max iters       & 150           & {300 ($\rho{\ge}0.6$) / 200} \\
Early stop      & {$<$0.5\% rel.\,$\Delta\mathcal{L}$ over 5 iters} & {val PPL pat.\,40 steps} \\
Val set         & ---           & {WikiText-2 val (${\sim}$145 seqs)} \\
Typical iters   & 15--25        & {40--100 (best restore)} \\
Precision       & fp32          & {bf16 fwd / fp32 params} \\
$\lambda_{\mathrm{la}}$ & 0.05 & --- \\
\bottomrule
\end{tabular}
\caption{Hyperparameters for L2 (block-level) and L3 (end-to-end).}
\label{tab:hyperparams}\label{app:hyperparams}
\end{minipage}

\vspace{12pt}
\noindent\textbf{Skip-L2 Training Dynamics.}\label{app:skip-l2-dynamics}
Figure~\ref{fig:skip-l2-dynamics} shows the full
training dynamics. Both paths use AdamW with cosine decay, early
stopping on WikiText-2 validation PPL, and best checkpoint restore.
At lr $5{\times}10^{-6}$, skip-L2 reaches a lower validation PPL
(11.98 vs.\ 12.07) but overfits past step 90. At lr
$2{\times}10^{-5}$, WikiText-2 test PPL is similar (11.4 vs.\ 11.9)
but PTB diverges by 24 points (62.8 vs.\ 86.9). L2's block-level
hidden-state matching provides ${\sim}$16M per-token constraints
(256 samples $\times$ 2048 positions $\times$ 32 blocks) that
regularize the factored parameters before L3 fine-tuning.
\end{table*}

\section{Robustness and Diagnostic Results}

\begin{figure}[t]
\centering
\includegraphics[width=\columnwidth]{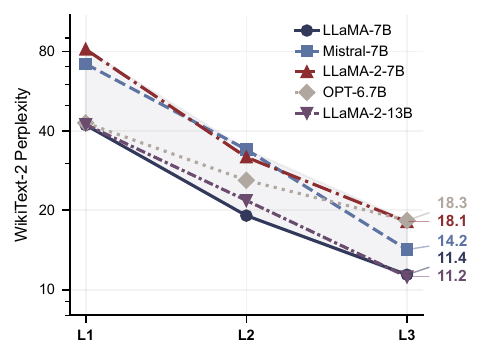}
\caption{WikiText-2 test perplexity across optimization levels
(L1$\to$L2$\to$L3) for five models (7B--13B) at $\rho{=}0.6$. 
The chain generalizes across architectures, though per-level gains are model-dependent.}
\label{fig:cross-arch}
\end{figure}

\begin{figure*}[t]
\centering
\includegraphics[width=\textwidth]{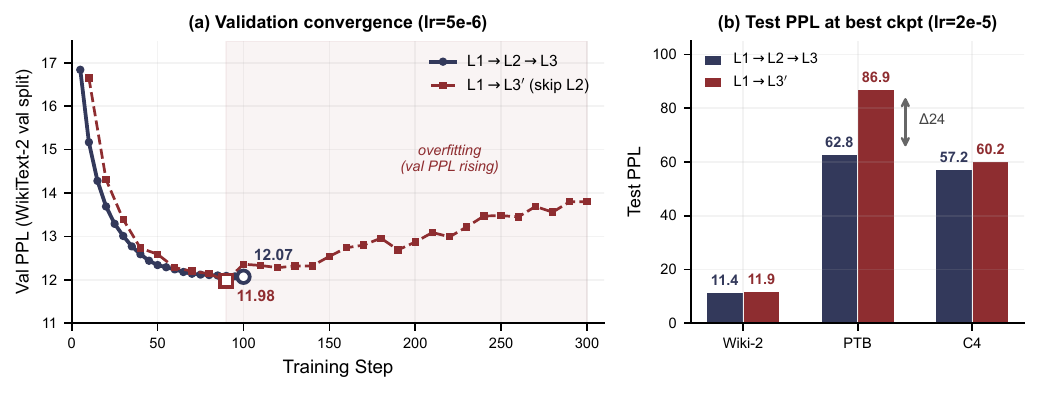}
\caption{Skip-L2 ablation on LLaMA-7B at $\rho{=}0.6$.
\textbf{(a)}~Validation PPL (lr $5{\times}10^{-6}$) converges similarly, but skip-L2 overfits past step 90.
\textbf{(b)}~Test PPL at best checkpoint (lr $2{\times}10^{-5}$): in-distribution (WikiText-2) matches, but OOD (PTB) diverges by 24 points.}
\label{fig:skip-l2-dynamics}
\end{figure*}

\label{app:robustness}

\paragraph{End-to-End Pipeline Seeds.}
Table~\ref{tab:seeds} repeats the full-versus-skip comparison across
three end-to-end pipeline seeds on LLaMA-7B at $\rho{=}0.6$. Each
seed controls L1 calibration sampling, L2 randomness and L3
randomness; within a seed the two arms share L1 factors, the
256-sequence pool, rank allocation, the L3 seed and schedule, and the
WikiText-2 selection rule, so the designed difference is the presence
of L2. Early stopping sets each run's length, and WikiText-2
validation alone selects checkpoints---PTB and C4 never train, tune
or select. The full chain is lower on all three corpora for every
seed. This resamples the whole pipeline rather than isolating
calibration-subset effects, and three seeds show a consistent
direction with variable effect size rather than low variance or
arbitrary-seed robustness.

\begin{table}[h]
\centering
\small
\setlength{\tabcolsep}{3pt}
\begin{tabular}{@{}lcc@{}}
\toprule
Seed & Full chain & Skip-L2 \\
\midrule
42   & 11.41 / 67.64 / 48.53 & 11.97 / \phantom{0}91.44 / 59.71 \\
123  & 11.46 / 72.76 / 54.21 & 12.14 / 107.90 / 63.67 \\
456  & 11.10 / 69.62 / 51.75 & 12.33 / \phantom{0}94.85 / 60.02 \\
\midrule
Mean & 11.32 / 70.01 / 51.50 & 12.14 / \phantom{0}98.06 / 61.13 \\
\bottomrule
\end{tabular}
\caption{End-to-end pipeline-seed check (LLaMA-7B, $\rho{=}0.6$),
reported as WikiText-2 / PTB / C4 perplexity.}
\label{tab:seeds}
\end{table}

\paragraph{Level-Wise Retention Diagnostics.}
Table~\ref{tab:levelwise} reports the seven-task zero-shot average
at each level. L2 improves all seven tasks over L1; L3 improves six
of seven over L2, reducing MathQA by $0.40$ points. All compressed
states remain far below the dense model, so we present these as
retention diagnostics rather than evidence of capability
preservation. These averages are computed from our own
\texttt{lm-evaluation-harness} v0.4 run over the canonical seed-42
states; Table~\ref{tab:svd-comparison}'s uncompressed row is quoted
from the baseline source~\cite{hu2026saes}, so the two dense figures
are not directly comparable and this table is not a decomposition of
that table's rows.

\begin{table}[h]
\centering
\small
\begin{tabular}{@{}lc@{}}
\toprule
State & Seven-task avg.\ (\%) $\uparrow$ \\
\midrule
Dense & 55.00 \\
L1    & 31.00 \\
L2    & 35.56 \\
L3    & 37.46 \\
\bottomrule
\end{tabular}
\caption{Level-wise zero-shot retention on LLaMA-7B at
$\rho{=}0.6$, same seven tasks as
Table~\ref{tab:svd-comparison}, our harness run.}
\label{tab:levelwise}
\end{table}

\paragraph{Generation and Reasoning Diagnostics.}
Perplexity gains do not transfer to open-ended generation at this
ratio (Table~\ref{tab:generation}). On a held-out 200-example SAMSum
subset evaluated with no prompt truncation---where the prompt and
stopping rule were fixed using dense-model validation only, before
any compressed state was scored---L2 and L3 improve ROUGE-L over L1,
but every compressed state reaches the 128-token ceiling on all
examples and shows high repetition (repeated-4-gram rates of
$50.9\%$--$72.8\%$), whereas the dense model follows the delimiter on
$199/200$ examples. On GSM8K, all compressed states score $0\%$
strict 5-shot exact match against $9.17\%$ for dense and therefore
cannot distinguish the levels. This subset is not the full or
standard SAMSum benchmark, and these results do not support reliable
or broad generation, summarization, dialogue, or
instruction-following preservation at $60\%$ removal.

\begin{table}[h]
\centering
\footnotesize
\setlength{\tabcolsep}{2pt}
\begin{tabular}{@{}lrrrr@{}}
\toprule
Diagnostic & Dense & L1 & L2 & L3 \\
\midrule
SAMSum ROUGE-L F1$\times100$ & 37.18 & 2.86 & 10.90 & 16.90 \\
SAMSum 128-tok.\ stops       & 0.5\% & 100\% & 100\% & 100\% \\
GSM8K 5-shot EM (strict)     & 9.17\% & 0\% & 0\% & 0\% \\
\bottomrule
\end{tabular}
\caption{Generation and reasoning diagnostics on LLaMA-7B at
$\rho{=}0.6$. Held-out 200-example SAMSum subset; not the standard
benchmark.}
\label{tab:generation}
\end{table}

\paragraph{Removable Overhead in the Block Loop.}
A bounded implementation study indicates that part of the
block-level stage's cost is removable overhead rather than intrinsic.
Caching fixed teacher and probe targets, keeping the frozen probe and
next block resident, evaluating only the required top-128 probe rows
with a custom backward, and dropping an unused teacher-logit cache
leaves the objective unchanged---losses and hidden-state gradients
agree with the reference implementation within $3\times10^{-6}$ and
$4\times10^{-5}$, and all three same-device pairs record the same
selected one-block objective value. On a one-block, five-step
benchmark run sequentially on each of three physical GPUs, this gives
a $1.537\times$ geometric-mean speedup for the block-optimization
stage and $1.302\times$ for the complete one-block compression call,
with peak allocated memory rising from $41.409$ to $42.067$\,GiB.
These measurements do not establish a full-stage or
hardware-independent speedup. The full gradient sweep over all 256
sequences at every Adam step remains the dominant cost, so larger
gains would require reducing or accelerating that sweep without
degrading compression quality.

\end{document}